# A Fast and Accurate Unconstrained Face Detector

Shengcai Liao, *Member, IEEE*, Anil K. Jain, *Fellow, IEEE*, and Stan Z. Li, *Fellow, IEEE*

**Abstract**—We propose a method to address challenges in unconstrained face detection, such as arbitrary pose variations and occlusions. First, a new image feature called Normalized Pixel Difference (NPD) is proposed. NPD feature is computed as the difference to sum ratio between two pixel values, inspired by the Weber Fraction in experimental psychology. The new feature is scale invariant, bounded, and is able to reconstruct the original image. Second, we propose a deep quadratic tree to learn the optimal subset of NPD features and their combinations, so that complex face manifolds can be partitioned by the learned rules. This way, only a single soft-cascade classifier is needed to handle unconstrained face detection. Furthermore, we show that the NPD features can be efficiently obtained from a look up table, and the detection template can be easily scaled, making the proposed face detector very fast. Experimental results on three public face datasets (FDDB, GENKI, and CMU-MIT) show that the proposed method achieves state-of-the-art performance in detecting unconstrained faces with arbitrary pose variations and occlusions in cluttered scenes.

## 1 Introduction

The objective of face detection is to find and locate faces in an image. It is the first step in automatic face recognition applications. Face detection has been well studied for frontal and near frontal faces. The Viola and Jones' face detector [1] is the most well known face detection algorithm, which is based on Haar-like features and cascade AdaBoost [2] classifier. However, in unconstrained scenes such as faces in a crowd, state-of-the-art face detectors fail to perform well due to large pose variations, illumination variations, occlusions, expression variations, out-of-focus blur, and low image resolution. For example, the Viola-Jones face detector fails to detect most of the face images in the Face Detection Data set and Benchmark (FDDB) database [3] (examples shown in Fig. 1) due to the difficulties mentioned above. In this paper, we refer to face detection with arbitrary facial variations as the unconstrained face detection problem. We are interested in face detection in unconstrained scenarios such as video surveillance or images captured by hand-held devices.

Numerous face detection methods have been developed following Viola and Jones' work [1], mainly focusing on extracting different types of features and developing different cascade structures. A variety of complex features [4], [5], [6], [7], [8], [9], [10], [11], [12], [13] have been proposed to replace the Haar-like features used in [1]. While these methods can improve the face detection performance to some extent, they generate a very large number (hundreds of thousands) of features and the resulting systems take too much time to train. Another development in face detection has been to learn different cascade structures for multiview face detection, such as parallel cascade [14], pyramid architecture [15], and Width-First-Search (WFS) tree [16]. All these methods need to learn one cascade classifier for each specific facial view (or view range). In unconstrained scenarios, however, it is not easy to define all possible views of a face, and the computational cost increases with an increasing number of classifiers in complex cascade structures. Moreover, these approaches require manual labeling of face pose in each training image.

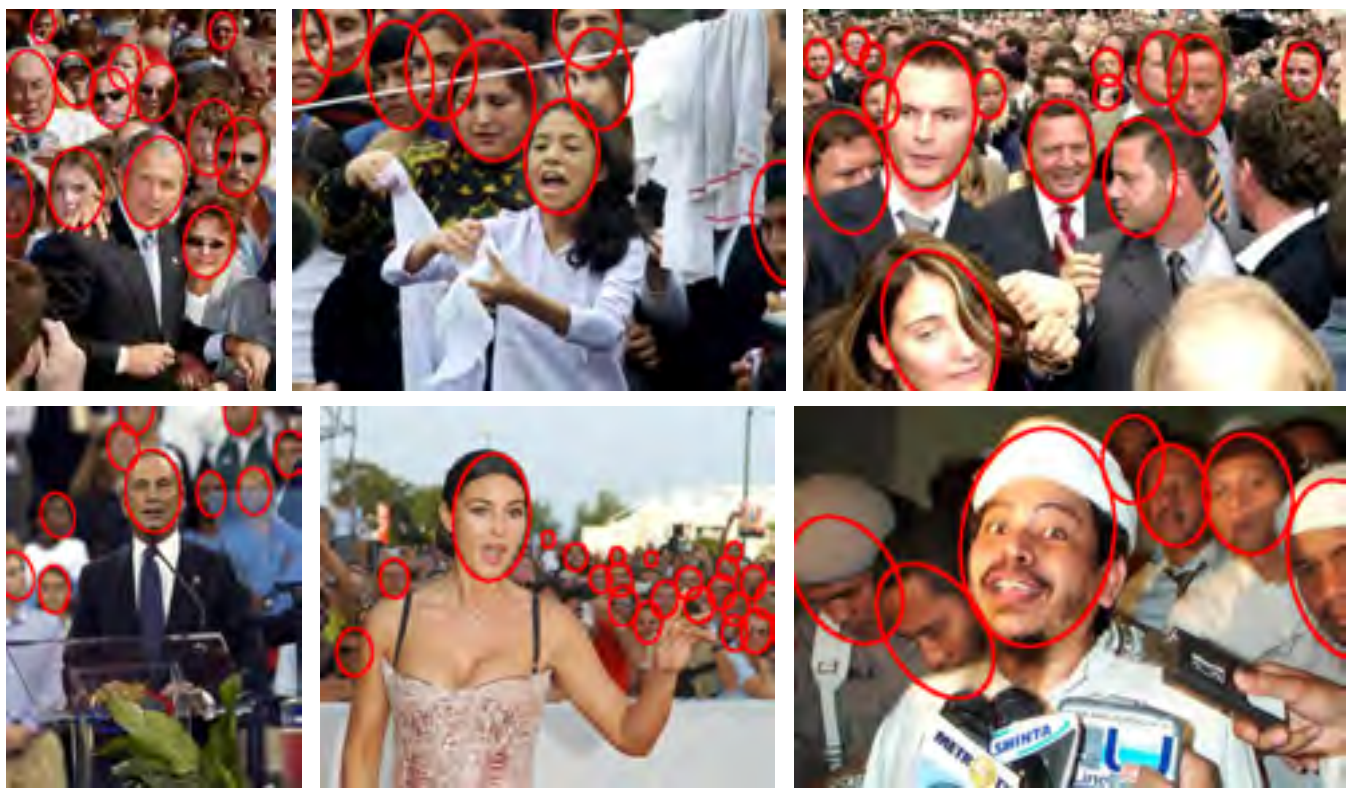

Fig. 1. Face images annotated (red ellipses) in the FDDB database [3].

While some of the available methods [14], [15], [16] can handle multiview faces, they are not able to simultaneously consider other challenges such as occlusion. In fact, since these methods require partitioning multiview data into known poses, occlusion is not easy to handle in this way. On the other hand, while several studies addressed face detection under occlusion [17], [18], [19], [20], [21], they constrained themselves to detect only frontal faces under occlusion. As discussed in [22], a robust face detection algorithm should be effective under arbitrary variations in pose and occlusion, which remains an unresolved challenging problem.

*Shengcai Liao and Stan Z. Li are with the National Laboratory of Pattern Recognition and the Center for Biometrics and Security Research, Institute of Automation, Chinese Academy of Sciences, Beijing 100190, China. E-mail: {scliao,szli}@nlpr.ia.ac.cn. Anil K. Jain is with the Dept. of Computer Science and Engineering, Michigan State University, East Lansing, MI 48824 USA. E-mail: jain@cse.msu.edu.* *This work was supported by NSFC #61203267, National Science and Technology Support Program #2013BAK02B01, and CAS Project #KGZD-EW-102-2.*

In this paper, we are interested in developing effective features and robust classifiers for unconstrained face detection with arbitrary facial variation. First, we propose a simple pixel-level feature, called the Normalized Pixel Difference (NPD). An NPD is computed as the ratio of the difference between any two pixel intensity values to the sum of their values, in the same form as the Weber Fraction in experimental psychology [23]. The NPD feature has several desirable properties, such as scale invariance, boundedness, and ability to reconstruct the original image. we further show that NPD features can be obtained from a look up table, and the resulting face detection template can be easily scaled for multiscale face detection.

Secondly, we propose a deep quadratic tree learning method and construct a single soft-cascade AdaBoost classifier to handle complex face manifolds and arbitrary pose and occlusion conditions. While individual NPD features may have "weak" discriminative ability, our work indicates that a subset of NPD features can be optimally learned and combined to construct more discriminative features in a deep quadratic tree. In this way, different types of faces can be automatically divided into different leaves of a tree classifier, and the complex face manifold in a high dimensional space can be partitioned in the learning process. This is the "divide and conquer" strategy to tackle unconstrained face detection in a single classifier, without pre-labeling of views in the training set of face images. The resulting face detector is robust to variations in pose, occlusion, and illumination, as well as to blur and low image resolution.

The novelty of this work is summarized as follows:

- A new type of feature, called NPD is proposed, which is efficient to compute and has several desirable properties, including scale invariance, boundedness, and enabling reconstruction of the original image.
- A deep quadratic tree learner is proposed to learn and combine an optimal subset of NPD features to boost their discriminability. In this way, only a single soft-cascade AdaBoost classifier is needed to handle unconstrained faces with occlusions and arbitrary viewpoints, without pose labeling or clustering in the training stage.

The advantages of the proposed approach include:

- The NPD feature evaluation is extremely fast, requiring a single memory access using a look up table.
- Multiscale face detection can be easily achieved by applying pre-scaled detection templates.
- The unconstrained face detector does not depend on pose specific cascade structure design; pose labeling or clustering in the training stage is also not required.
- The face detector is able to handle illumination variations, pose variations, occlusions, out-of-focus blur, and low resolution face images in unconstrained scenarios.

The source code of the proposed method is available in http://www.cbsr.ia.ac.cn/users/scliao/projects/npdface/. The remainder of this paper is organized as follows. In Section 2 we review the related work. In Section 3 we introduce the NPD feature space. The proposed NPD based face detection method is presented in Section 4. Experimental results are provided in Section 5. Finally, we summarize the contributions in Section 6.

## 2 RELATED WORK

As indicated in a survey of face detection methods [24], the most popular face detection methods are appearance based, which use local feature representation and classifier learning. Viola and Jones' face detector [1] was the first one to apply rectangular Haar-like features in a cascaded AdaBoost classifier for real-time face detection. Many approaches have been proposed around the Viola-Jones detector to advance the state of the art in face detection. Lienhart and Maydt [4] proposed an extended set of Haar-like features, where $45^\circ$ rotated rectangular features were introduced. Li et al. [5] proposed another extension of Haar-like features, where the rectangles can be spatially set apart with a flexible distance. A similar feature, called the diagonal filter was also proposed by Jones and Viola [6]. Various other local texture features have been introduced for face detection, such as the modified census transform [7], local binary pattern (LBP) [8], MB-LBP [11], LBP histogram [10], and the locally assembled binary feature [12]. These features have been shown to be robust to illumination variations. Mita et al. [9] proposed the joint Haar-like features to capture the co-occurrence of effective Haar-like features. Huang et al. [16] proposed a sparse feature set in a granular space, where granules were represented by rectangles, and each individual sparse feature was learned as a combination of granules. A problem with the approaches in [9] and [16] is that the joint feature space is very large, making the optimal combination a difficult task.

While more sophisticated features may provide better discrimination power than Haar-like features for the face detection task, they generally increase the computational cost. In contrast, ordinal relationships among image regions are simple yet effective image features [25], [26], [27], [28], [29], [30], [31]. Sinha [25] studied several robust ordinal relationships in face images and developed a face detection method accordingly. Liao et al. [28] further showed that ordinal features can be effectively learned by AdaBoost classifier for face recognition. Sadr et al. [26] showed that pixelwise ordinal features (POF), i.e. ordinal relationship ($x > y$) between any two pixels, can faithfully encode image structures. Lepetit and Fua [29] applied POF features in random trees for keypoint recognition. Shotton [32] applied POF features in random forests for image categorization and segmentation. For facial analysis, Baluja et al. [27] showed that POF features are good enough for discriminating between five facial orientations, a relatively simpler task than face detection. Wang et al. [31] applied the random forest classifier together with POF features for facial landmark localization. Abramson and Steux [30] proposed a pixel control point based feature for face detection, where each feature is associated with two sets of pixel locations (control points).

Besides different feature representations, some researchers have also tried different AdaBoost algorithms and weak classifiers. For weak classifiers utilized in boosting, Lienhart et al. [33] and Brubaker et al. [34] have shown that classification and regression trees (CART) [35] work better than simple decision stumps. In this paper, we show that the optimal ordinal/contrastive features and their combinations can be learned by integrating the proposed NPD features in a deep quadratic tree. In this way, unconstrained face variations can be automatically partitioned into different leaves of the learned quadratic tree classifier.

Given that the original Viola-Jones face detector has limitations for multiview face detection [24], various cascade structures have been proposed to tackle multiview face detection [6], [14], [15], [16]. Jones and Viola [6] extended their face detector by training one face detector for each specific pose. To avoid evaluating all face detectors on each scanning subwindow, they developed a pose estimation step (similar to Rowley et al. [36]) before face detection, and then only the face detector trained on that estimated pose was applied. In this two-stage detection structure, if the pose estimation is not reliable, the face is not likely to be detected in the second stage. Wu et al. [14] proposed a parallel cascade structure for multiview face detection, where all face detectors tuned to different views have to be evaluated for each scanning window; they did use the first few cascade layers of all face detectors to estimate the pose for speedup. Li and Zhang [15] proposed a coarse-to-fine pyramid architecture for multiview face detection, where the entire range of face poses was divided into increasingly smaller subranges, resulting in a more efficient detection structure. Huang et al. proposed a WFS tree based multiview face detection approach, which also works in a coarse-to-fine manner. They proposed the Vector Boost algorithm for multiclass learning, which is well suited for multiview pose estimation. However, all these methods need to learn a cascade classifier for each specific view (or view range) of a face, requiring an input face image to go through different branches of the detection structure. Hence, their computational cost generally increases with the number of classifiers in complex cascade structures. Moreover, these approaches require manual labeling of the face pose in each training image.

Instead of designing a detection structure, Lin and Liu [19] proposed to learn the multiview face detector as a single cascade classifier. They derived a multiclass boosting algorithm, called MBHBoost by sharing features among different classes. This is a simpler approach to multiview face detection than designing complex cascade structures. Nevertheless, it still requires manual labeling of poses. In uncontrolled environments, however, it is not easy to define specific views of a face by discretizing the pose space, because a face could be in arbitrary pose simultaneously in yaw (out-of-plane), roll (in-plane), and pitch (up-and-down) angles. To avoid manual labeling, Seemann et al. [37] suggested learning viewpoint clusters automatically for object detection. However, for human faces, Kim and Cipolla [38] showed that clustering by traditional techniques like K-Means does not result in categorized poses. They hence proposed a multi-classifier boosting (MCBoost) for human perceptual clustering of object images, which showed promise for clustering face poses. However, the clusters are not always related to pose variations; in addition to different pose clusters, they also obtained clusters with various illumination variations.

Face detection in presence of occlusion is also an important issue in unconstrained face detection, but it has received less attention compared to multiview face detection. This is probably because, compared to pose variations, it is more difficult to categorize arbitrary occlusions into predefined classes. Hotta [17] proposed a local kernel based SVM method for face detection, which was better than global kernel based SVM in detecting occluded frontal faces. Lin et al. [18] considered 8 kinds of manually defined facial occlusions by training 8 additional cascade classifiers besides the standard face detector. Lin and Liu [19] further proposed the MBHBoost algorithm to handle faces with one of 12 in-plane rotations or one of 8 types of occlusions, with each kind of rotation and occlusion treated as a different class. Chen et al. [20] proposed a modified Viola-Jones face detector, where the trained detector was divided into sub-classifiers related to several predefined local patches, and the outputs of sub-classifiers were fused. Goldmann et al. [21] proposed a component-based approach for face detection, where the two eyes, nose, and mouth were detected separately, and further connected in a topology graph. However, none of the above methods considered face detection with both occlusions and pose variations simultaneously in unconstrained scenarios. As discussed in [22], a robust face detector should be effective under arbitrary variations in pose and occlusion, which has not yet been solved.

Recently, unconstrained face detection has gained attention. Jain and Learned-Miller [3] developed the FDDB database and benchmark for the development of unconstrained face detection algorithms. This database contains images collected from the Internet, and presents challenging scenarios for face detection. Subburaman and Marcel [39] proposed a fast bounding box estimation technique for face detection, where the bounding box is predicted by small patch based local search. Jain and Learned-Miller [40] proposed an online domain adaption approach to improve the performance of the Viola-Jones face detector on the FDDB database. Li et al. [13] proposed the use of SURF feature [41] in an AdaBoost cascade, and area under the curve (AUC) criterion to speed up the face detector training. Shen et al. [42] proposed an exemplar-based face detection approach, which retrieves images from a large annotated face dataset; facial landmark locations are inferred from the annotations. This method is further improved in [43] by boosting. Li et al. [44] proposed a probabilistic elastic part (PEP) model to adapt any pre-trained face detector to a specific image collection like FDDB by an additional post-processing classifier. Zhu and Ramanan [45] proposed to jointly detect a face, estimate its pose, and localize face landmarks in the wild by a Deformable Parts-based Model (DPM), which was further improved in [46] and [47]. Chen et al. [48] proposed to combine the face detection and landmark estimation tasks in a joint cascade framework to refine face detection by precise landmark detections. Yang et al. [49] investigated the use of channel features for face detection, which achieves promising performance. Despite the availability of these methods for unconstrained face detection, the detection accuracy is still not satisfactory, especially when the detector is required to have low false alarms.

# 3 Normalized Pixel Difference Feature Space

The Normalized Pixel Difference (NPD) feature between two pixels in an image is defined as

$$f(x, y) = \frac{x - y}{x + y}, \quad (1)$$

where $x, y \geq 0$ are intensity values of the two pixels[1], and $f(0, 0)$ is defined as 0 when $x = y = 0$.

1. For ease of representation, sometimes we also denote $x$ and $y$ as pixels instead of pixel values. We use subscripts to differentiate between pixel and pixel values only when pixel locations are under discussion.

The NPD feature measures the relative difference between two pixel values. The sign of $f(x, y)$ indicates the ordinal relationship between the two pixels $x$ and $y$, and the magnitude of $f(x, y)$ measures the relative difference (as a percentage of the joint intensity $x + y$) between $x$ and $y$. Note that the definition $f(0, 0) \triangleq 0$ is reasonable because, in this case, there is no difference between the two pixels $x$ and $y$. Compared to the absolute difference $|x - y|$, NPD is invariant to scale change of the pixel intensities.

Weber, a pioneer in experimental psychology, stated that the just-noticeable difference in the magnitude change of a stimulus is proportional to the magnitude of the stimulus, rather than its absolute value [23]. This is known as the Weber's Law. In other words, the human perception of difference in stimulus is often measured as a fraction of the original stimulus, that is, in a form $\Delta I/I$, which is called the Weber Fraction. Chen et al. [50] proposed a local image descriptor, called Weber's Law Descriptor for face recognition, which was computed from Weber Fractions of pixels in a $3 \times 3$ window. The proposed feature in Eq. (1) has also been used in other fields such as remote sensing, where the Normalized Difference Vegetation Index (NDVI) [51] is defined as the difference to sum ratio between the visible red and the near infrared spectra to estimate the green vegetation coverage.

The NPD feature has a number of desirable properties. First, the NPD feature is antisymmetric, so either $f(x, y)$ or $f(y, x)$ is adequate for feature representation, resulting in a reduced feature space. Therefore, in an $s \times s$ image patch (vectorized as $p \times 1$, where $p = s \cdot s$), NPD feature $f(x_i, x_j)$ for pixel pairs $1 \leq i < j \leq p$ is computed, resulting in $d = p(p-1)/2$ features. For example, in a $20 \times 20$ face template, there are $(20 \times 20) \times (20 \times 20 - 1)/2 = 79,800$ NPD features in total. We call the resulting feature space the NPD feature space, denoted as $\Omega_{npd}$ ($\in \mathbb{R}^d$).

Second, the sign of $f(x, y)$ is an indicator of the ordinal relationship between $x$ and $y$. Ordinal relationship has been shown to be an effective encoding for object detection and recognition [25], [26], [28] because ordinal relationship encodes the intrinsic structure of an object image and it is invariant under various illumination changes [25]. However, simply using the sign to encode the ordinal relationship is likely to be sensitive to noise when $x$ and $y$ have similar values. In the next section we will show how to learn robust ordinal/contrastive relationships with NPD features.

Third, the NPD feature is scale invariant, which is expected to be robust against illumination changes. This is important for image representation, since illumination change is always a troublesome issue for both object detection and recognition.

Fourth, as shown in Appendix A, the NPD feature f(x,y) is bounded in [-1,1]. The bounded property makes the NPD feature amenable to histogram binning or threshold learning in tree-based classifiers [1].

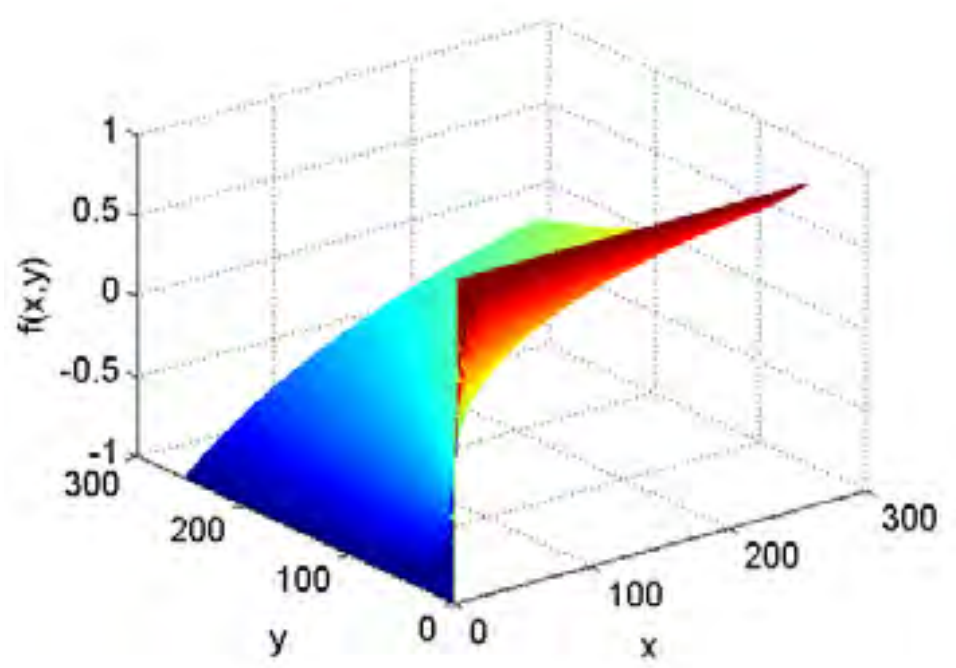

Fig. 2. A plot of the NPD function $f(x, y)$.

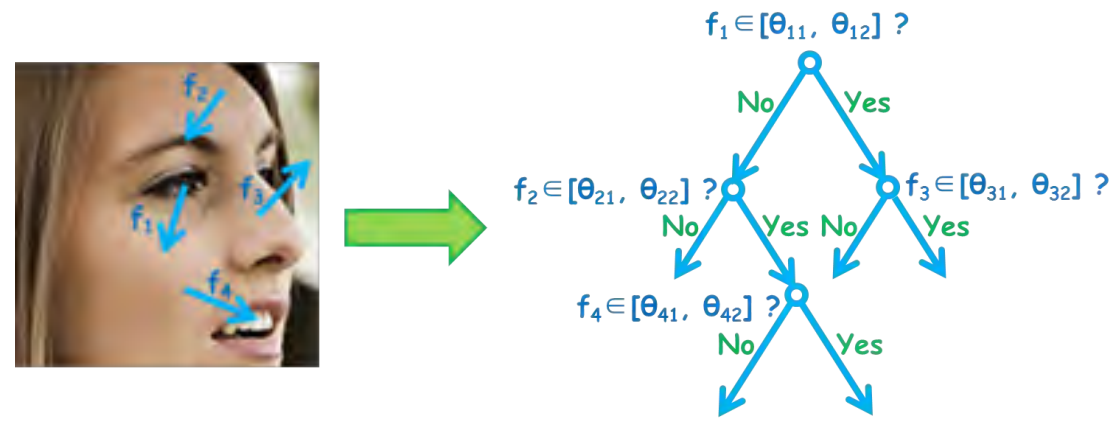

Fig. 3. Learning and combining NPD features in a deep quadratic tree. Left: four NPD features are automatically selected in the learning process. Right: the four features are optimally combined in a deep quadratic tree for face/nonface prediction.

Fig. 2 shows that $f(x, y)$ is a bounded function and it defines a nonlinear surface.

***Theorem 1 (Reconstruction):*** Given the NPD feature vector $\mathbf{f} = (f(x_1, x_2), f(x_1, x_3), \ldots, f(x_{p-1}, x_p))^T \in \Omega_{npd}$, the original image $I = (x_1, x_2, \ldots, x_p)^T$ can be reconstructed up to a scale factor.

The proof of Theorem 1 is shown in Appendix B, which also gives a linear-time approach to reconstruct the original image up to a scale factor. Theorem 1 states that each point in the feature space $\Omega_{npd}$ corresponds to a group of intensity-scaled images in the original pixel intensity space. In contrast, the scale invariance property says that all intensity-scaled images are "compressed" to a point in the bounded feature space $\Omega_{npd}$. Therefore, $\Omega_{npd}$ is a feature space which is invariant to scale variations, but it carries all the necessary information from the original space.

## 4 NPD FOR FACE DETECTION

### 4.1 Deep Quadratic Tree

The classic Viola-Jones face detector [1] learns representative features by boosted stumps. A stump is a basic tree classifier with one threshold that splits a node in two leaves. There are two limitations with stumps. First, this shallow structure cannot capture interactions between different feature dimensions. Second, the simple thresholding ignores higher-order information contained in a feature.

Therefore, in this paper, we consider a quadratic splitting strategy and a deeper tree structure. Specifically, for a feature $x$, we consider the tree node splitting as

$$(ax^2 + bx + c) < t, \tag{2}$$

where $a, b, c$ are constants w.r.t. $x$, and $t$ is the splitting threshold. With appropriate coefficients, this corresponds to checking whether $x$ is in a range $[\theta_1, \theta_2]$ or not, where $\theta_1$ and $\theta_2$ are two learned thresholds. Compared to the original linear splitting $x < t$, Eq. (2) considers both the first-order and second-order information of $x$, enabling a better interpretation of the splitting rule. Particularly, for the proposed NPD feature, three kinds of object structures can be learned:

$$-1 \leq \frac{x-y}{x+y} \leq \theta < 0, \tag{3}$$

$$0 < \theta \leq \frac{x-y}{x+y} \leq 1, \tag{4}$$

$$\theta_1 \leq \frac{x-y}{x+y} \leq \theta_2, \tag{5}$$

where $\theta_1 < 0$ and $\theta_2 > 0$. Eq. (3) applies if the object pixel $x$ is notably darker than pixel $y$ (e.g. $f_1$ in Fig. 3), while Eq. (4) covers the case when pixel $x$ is notably brighter than pixel $y$ (e.g. $f_2$ in Fig. 3). These two kinds of structures can also be learned by a classic stump. They are also known as ordinal relationships similar as in [25], except that a better threshold is learned instead of the default threshold 0. In contrast, if Eq. (5) does not hold, then there will be a notable edge or contrast between pixels $x$ and $y$ (e.g. $f_3$ and $f_4$ in Fig. 3), but the polarity is uncertain. For example, $f_3$ in Fig. 3 represents a notable edge between the face and background, but the background pixel can be either darker or brighter than the face. This kind of contrastive structure can only be learned by a quadratic splitting.

In practice, instead of solving Eq. (2) for quadratic splitting, we propose to quantize the feature range into L discrete bins (e.g. L=256 in this paper), and do an exhaustive search to determine the two optimal thresholds, where the weighted mean square error is applied as the optimal splitting criterion. Thanks to the bounded property of the proposed NPD feature, this quantization can be easily done. Besides, we build an L-bin histogram of the sample weights, and apply a one-dimensional integral technique similar as in [1] to speed up the splitting.

Furthermore, we apply the quadratic splitting to learn a deep tree (e.g. depth of eight in this paper), instead of a stump or a shallow tree for face detection. This way, several NPD features are optimally combined together to represent the intrinsic face structure. An example is shown in Fig. 3. The proposed deep quadratic tree is well suited for face detection with arbitrary pose variations, since similar views can be clustered in the same leaf node of the tree.

### 4.2 Face Detector

Given that the proposed NPD features contain redundant information, we also apply the AdaBoost algorithm to select the most discriminative features and construct strong classifiers [1]. We adopt the Gentle AdaBoost algorithm [2] to learn the NPD feature based deep quadratic trees.

As in [1], a cascade classifier is further learned for rapid face detection. We only learn one single cascade classifier for unconstrained face detection robust to occlusions and pose variations. This implementation has the advantage that there is no need to label the pose of each face image manually or cluster the poses before training the detector. In the learning process, the algorithm automatically divides the whole face manifold into several sub-manifolds by the deep quadratic trees. Besides, we adopt the soft cascade structure [52] for efficient training and early rejection of negative samples. Specifically, soft cascade can be regarded as a single AdaBoost classifier with one exit per weak classifier. In each iteration, a deep quadratic tree is learned as the weak classifier, and a threshold of the current AdaBoost classifier is also learned for rejecting nonfaces. Finally, the learned deep quadratic trees and thresholds are aggregated sequentially to represent an ensemble [2].

Below is a summary of how the proposed method handles the unconstrained face detection problem.

- **Pose**. Pose variations are handled by learning NPD features in boosted deep quadratic trees, where different views can be automatically partitioned into different leaves of the trees.
- **Occlusion**. In contrast to Haar-like features that are sensitive to occlusions because of large support [18], NPD features are computed by only two pixel values, making them robust to occlusion.
- **Illumination**. Since NPD features are scale invariant, they are robust to illumination changes.
- **Blur or low image resolution**. Because the NPD features involve only two pixel values, they do not require rich texture information on the face. This makes NPD features effective in handling blurred or low resolution face images.

#### 4.2.1 Implementation Details

we used the Annotated Facial Landmarks in the Wild (AFLW) database [53] for training our unconstrained face detector. The AFLW database contains 25,993 face annotations in 21,997 real-world images collected from Flickr. This is an unconstrained face database including large face variations in pose, illumination, expression, ethnicity, age, gender, etc. We cropped 21,730 face images from AFLW. Together with their mirrored images and perturbations in positions, we had 217,300 face images in total for training. Some examples are shown in Fig. 4 (left). For bootstrapping nonface images, we also used the AFLW images, but masked the facial regions with random images containing no faces, as shown in Fig. 4 (right). We used a detection template of $24 \times 24$ pixels. We set the maximum depth of the tree classifiers to be learned as eight, so that at most eight NPD features need to be evaluated for each tree classifier. In the soft cascade training, we set the threshold of each exit as the minimal score of positive samples, i.e. we did not reject positive samples during training. Our final detector contains 1,226 deep quadratic trees, and 46,401 NPD features. Nevertheless, the average number of feature evaluations per detection window is only 114.5 considering stagewise nonface rejection, which is quite reasonable.

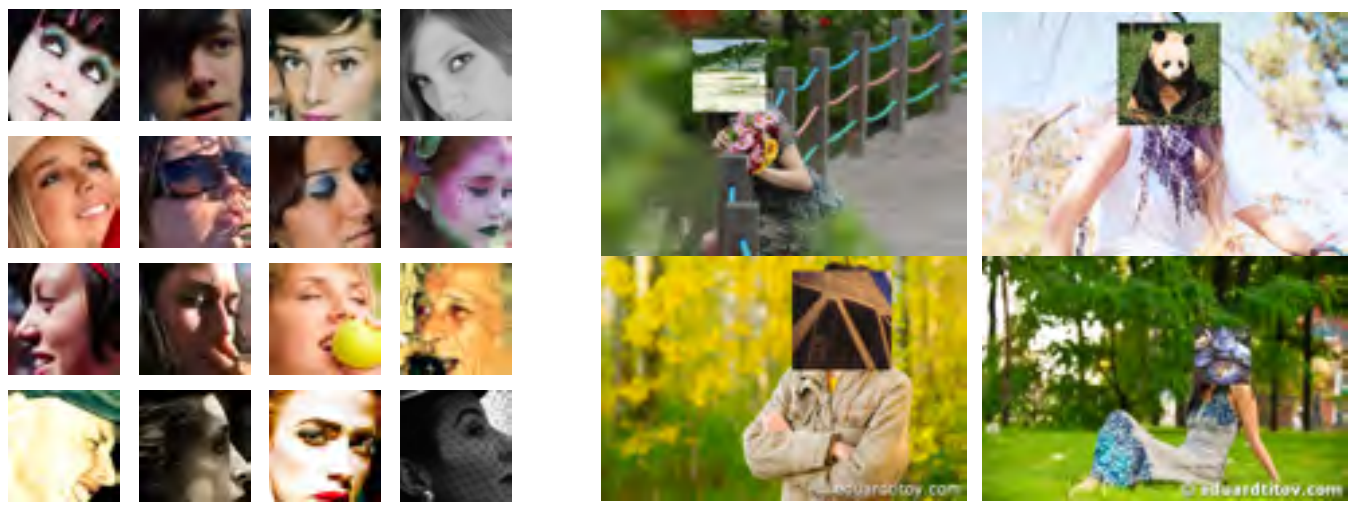

Fig. 4. Example face (left) and nonface (right) images from AFLW [53] for face detector training.

For an analysis, we also trained a near frontal face detector using the proposed NPD features and the classic cascade of regression trees (CART [35]) with depth of four. A subset of the training data[2] in [13] was used, including 12,102 face images and 12,315 nonface images. The detection template is $20 \times 20$ pixels. The detector cascade contains 15 stages, and for each stage, the target false accept rate was 0.5, with a detection rate of 0.999.

#### 4.2.2 Detector Speed Up

To further speed up the learned NPD detector for face detection, we develop the following two techniques. First, for 8-bit gray images, we build a $256 \times 256$ look up table to store pre-computed NPD features. This way, computing $f(x, y)$ in Eq. 1 only requires one memory access from the look up table.

Second, the learned face detection template (e.g. $20 \times 20$ used in this paper) can be easily scaled to enable multiscale face detection. So, we pre-compute multiscale detection templates and apply them to detect faces at various scales. This way, iterative rescaling of images for multiscale detection is avoided.

## 5 Experiments

We evaluate the performance of the NPD face detector on three public-domain databases, FDDB [3],

2. https://sites.google.com/site/leeplus/publications/facedetectionusingsurfcascade

GENKI [54], and CMU-MIT [36]. We also provide an analysis of the proposed method, report the face detection speed, and report unconstrained face detection performances under illumination variations, pose variations, occlusion, and blur, respectively.

In the test stage, a scale factor of 1.2 was set for multiscale detection. A postprocessing method similar to the OpenCV face detection module was implemented, which merges nearby detections by the disjoint set algorithm. For each detected face, we summarized the scores of AdaBoost classifiers in all stages of the cascade to be the final score; this score was used to generate the Receiver Operating Characteristic (ROC) curves.

### 5.1 Evaluation on FDDB Database

The FDDB dataset [3] covers challenging scenarios for face detection. Images in FDDB comes from the Faces in the Wild dataset [55], which is a large collection of Internet images collected from the Yahoo News. It contains 2,845 images with a total of 5,171 faces, with a wide range of challenging scenarios including arbitrary pose, occlusions, different lightings, expressions, low resolutions, and out-of-focus faces. All faces in the database have been annotated with elliptical regions. Fig. 1 shows some examples of the annotated faces from the FDDB database.

For benchmark evaluation, Jain and Learned-Miller [3] provided an evaluation code for a comparison of different face detection algorithms. There are two metrics for performance evaluation based on ROC: discrete score metric and continuous score metric, which correspond to coarse match (similar to previous evaluations in the face detection literature) and precise match, respectively, between the detection and the ground truth. The database is divided into 10 subsets for performance evaluation, and the obtained detection results are accumulated to generate the ROC curve[3].

We compared our method with state-of-the-art results reported on the FDDB website[4]. Table 1 shows a comprehensive comparison of detection rates of various algorithms on the FDDB database at FP=0, 10, and 100, where methods marked with a ☆ were trained on the same AFLW database [53] as ours. It can be observed that the proposed method outperforms most of the baseline methods except four methods [46], [47], [48], [49] published recently. The proposed NPD face detector is the second best one at FP=0 for the discrete metric and the third best one for the continuous metric. Specifically, the NPD detector detects about 54% of the annotated FDDB faces in coarse sense (50% overlap with ground truth) without any false alarms.

TABLE 1
Comparison of detection rates (%) with both discrete and continuous metrics on the FDDB database [3]*

| | Discrete Metric | | | Continuous Metric | | |
|---|---|---|---|---|---|---|
| | FP = 0 | FP = 10 | FP = 100 | FP = 0 | FP = 10 | FP = 100 |
| ☆ ACF-Multiscale [49] | **58.21** | 75.38 | 81.67 | 43.27 | 55.83 | 60.43 |
| ☆ NPD | 54.15 | 72.31 | 77.97 | 40.64 | 53.93 | 58.04 |
| ☆ Boosted Examplar [43] | 52.47 | 69.29 | 80.82 | 37.07 | 48.75 | 56.87 |
| ☆ Yan-DPM [46] | 51.61 | 75.98 | 81.36 | **43.70** | 63.48 | 67.70 |
| ☆ HeadHunter [47] | 34.94 | 72.58 | 83.41 | 28.71 | 58.94 | 67.28 |
| Joint Cascade [48] | 33.44 | **78.84** | **83.91** | 29.84 | **69.15** | **73.06** |
| Zhu-Ramanan [45] | 27.38 | 63.88 | 73.08 | 21.25 | 48.62 | 55.40 |
| SURF-Multiview [56] | 12.40 | 69.43 | 80.60 | 8.49 | 46.82 | 54.37 |
| VJGPR [40] | 4.58 | 15.76 | 51.00 | 2.95 | 10.20 | 33.16 |
| Mikolajczyk et al. [57] [3] | 3.25 | 10.23 | 33.28 | 2.10 | 6.61 | 21.67 |
| Viola-Jones [1] [3] | 1.39 | 10.02 | 32.64 | 0.90 | 6.48 | 21.26 |
| ☆ XZJY [42] | 0.31 | 7.91 | 67.51 | 0.19 | 4.99 | 43.40 |
| ☆ Koestinger et al. [58] | 0.19 | 21.47 | 57.03 | 0.14 | 15.38 | 40.55 |
| Segui et al. [59] | 0.00 | 15.08 | 67.94 | 0.00 | 9.78 | 43.76 |
| PEP [44] | n/a | 8.43 | 73.35 | n/a | 5.38 | 47.30 |
| Subburaman-Marcel [39] | n/a | 0.54 | 17.25 | n/a | 0.36 | 11.27 |

* Red numbers represents the best results, while blue numbers are the second best results. Methods marked by ☆ used the same AFLW [53] training data.

The ROC curves of recent methods are depicted in Fig. 5 for the discrete score metric and in Fig. 6 for the continuous score metric. In both Figs. 5 and 6, the curve labels in the legend are sorted in descending order of the detection rates at zero false positives (FP=0). It can be observed that the proposed NPD detector is among the top performers for the discrete metric, though it is not as good as the four recent methods for the continuous metric. However, note that the FDDB database uses ellipses for ground-truth of face annotations, and several methods (e.g. Yan-DPM [46], and HeadHunter [47]) output similar elliptical detections to improve the performance especially with the continuous metric. The proposed detector outputs square detections, followed by a 20% horizontal expansion and 50% vertical expansion as suggested in [43]. This processing is not as good as making elliptical detections, but is still better than the original square detections.

Compared to recent methods, the Joint Cascade algorithm [48] is the most competitive one to us in terms of accuracy and speed (see Sec. 5.6). However, the Joint Cascade method used a sophisticated postprocessing classifier to remove hard negatives and hence improved the results. Other methods are not efficient as compared in Sec. 5.6. Especially, the DPM based methods [45], [46], [47] are known to be quite slow. The method of Zhu-Ramanan [45] has the advantage of learning from only hundreds of face images and it jointly outputs face bounding box, pose, and landmarks. But it requires manual landmark and pose annotations as face prior knowledge before train-

3. According to [3], ten ROC curves should be obtained and averaged for the final performance report, however, what is actually done in the FDDB results webpage (Footnote 4) is that all detection results of the 10 subsets are first merged, and then a single ROC curve is evaluated. We followed the latter one for consistency to existing results.

4. http://vis-www.cs.umass.edu/fddb/results.html

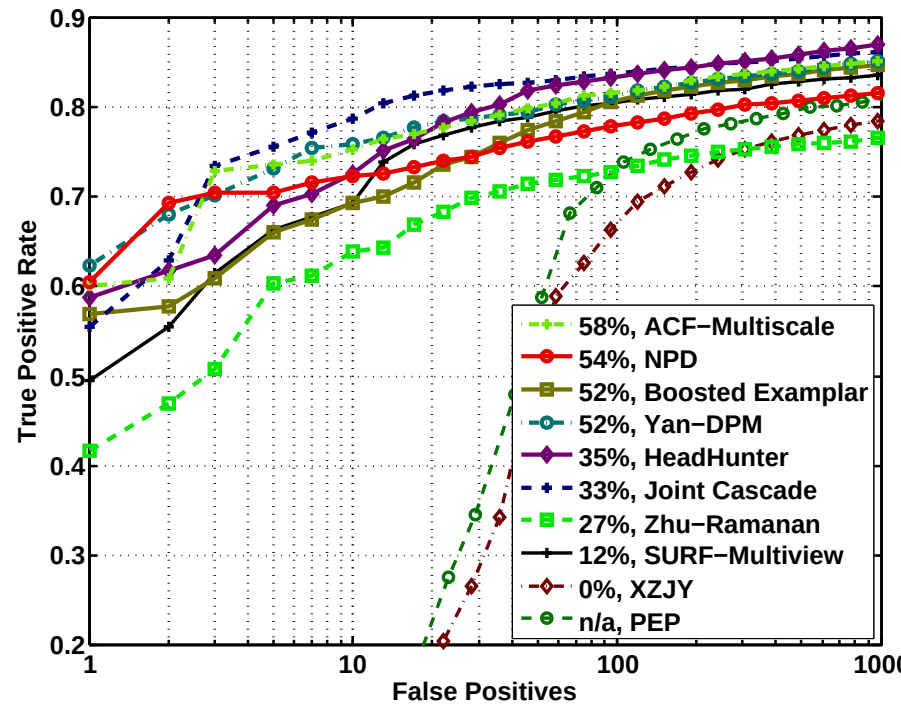

Fig. 5. ROC curves of recent methods on the FDDB database [3] with the discrete score metric.

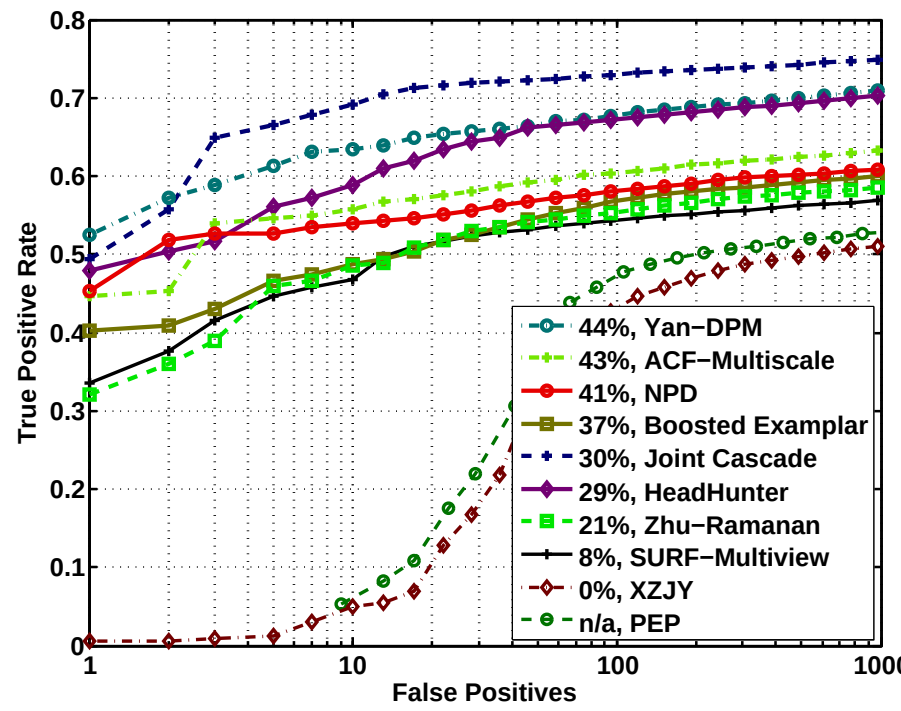

Fig. 6. ROC curves of recent methods on the FDDB database [3] with the continuous score metric.

ing. The performance of the Zhu-Ramanan model is quite impressive considering such a small training data. However, the runtime cost of their model is very expensive. As reported in [43], for a $1480 \times 986$ image, Zhu and Ramanan's detector takes 231 seconds to run and allocates up to 2GB memory. In contrast, our model is more efficient, requiring only a few milliseconds per image and only 50MB of memory as discussed in Sec. 5.6.

Fig. 7 shows some examples of detected faces in the FDDB database by the proposed NPD method. Many rotated, occluded, and out-of-focus faces can be successfully detected by the proposed method. Some faces (e.g. the 2nd image in row 1, and the 4th image in row 3 in Fig. 7) that are not annotated in the ground truth can still be detected by the proposed method. However, there are a number of faces that cannot be detected by the proposed method, especially in very crowded scenes (see the 1st image and the 3rd image in row 1, and the last two images in row 5 in Fig. 7). Therefore, unconstrained face detection in crowded scenes is still very challenging and deserves more attention.

### 5.2 Evaluation on GENKI Database

The GENKI database [54] was collected by the Machine Perception Laboratory, University of California, San Diego. We evaluated the current release of the GENKI database, GENKI-R2009a, on its SZSL subset, which contains 3,500 images collected from the Internet. These images include a wide range of backgrounds, illumination conditions, geographical locations, personal identity, and ethnicity. Some examples of face images from the GENKI database are shown in Fig. 9, with labeled detections by the proposed NPD method. Most images in the GENKI dataset contain only a single face. In that sense, the GENKI dataset is not as challenging as the FDDB dataset. Some of the images in the GENKI-SZSL dataset contain faces that are not labeled, therefore they are not suitable for the face detection evaluation task. After removing such unlabeled images, we are left with 3,270 images for face detection evaluation. We evaluated our unconstrained face detector, as well as the Viola-Jones face detector implemented in OpenCV 2.4, and a commercial face detector, PittPatt [60]. We again used the benchmark evaluation code in [3] for performance evaluation, but slightly modified the code for allowing ground truth annotations as rectangles. The ROC curves of the three methods are shown in Fig. 8 for both the discrete and continuous score metrics. The results show that the proposed NPD face detector significantly outperforms both the Viola-Jones and PittPatt face detectors.

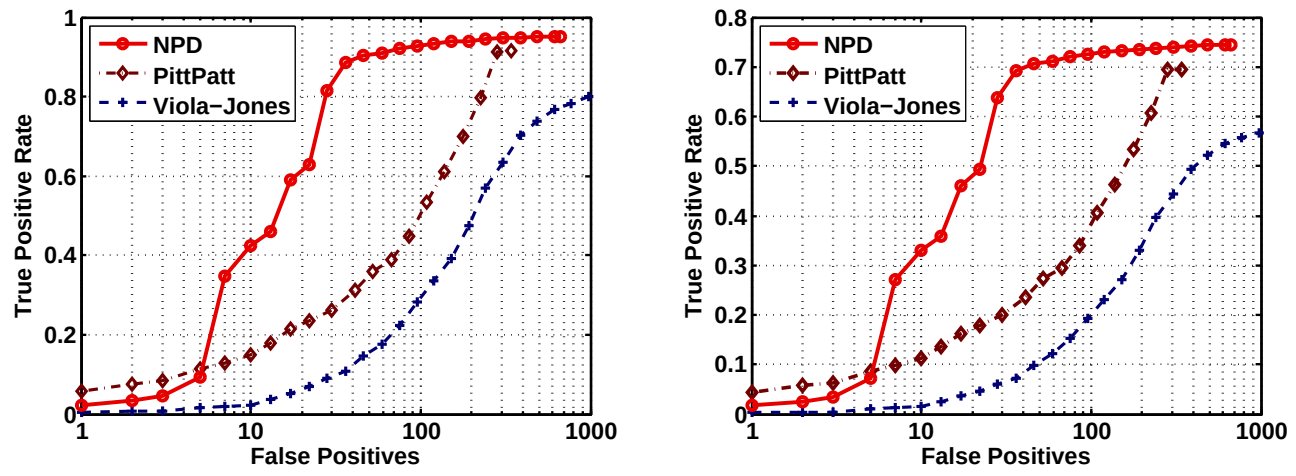

(a) discrete (b) continuous

Fig. 8. ROC curves for face detection on the GENKI-SZSL dataset [54] with (a) discrete and (b) continuous score metrics.

### 5.3 Evaluation on CMU-MIT Database

The CMU-MIT face dataset [36] is one of the early benchmarks for face detection. The CMU-MIT frontal face data set contains 130 gray-scale images with a total of 511 faces, most of which are not occluded. We applied our frontal NPD face detector described in Subsection 4.2.1 on this database. We also used the modified benchmark evaluation code from [3] with the discrete score metric for performance evaluation. Fig. 10 shows the ROC curves for the proposed NPD face detector, the Soft cascade method [52], the SURF cascade method [13], and the Viola-Jones detector [1]. The results show that, compared to the Viola-Jones frontal face detector, the NPD detector performs better when the number of false positives, FP $< 50$, while it is slightly worse than Viola-Jones at higher FPs. Compared to the SURF cascade detector, the NPD detector is better when FP $< 3$, but SURF cascade method outperforms NPD at higher FPs. Note that the SURF cascade method uses a face template of size $40 \times 40$ pixels, which is four times larger than our face detection template ($20 \times 20$ pixels). Generally, a larger face template contains more features for face description, but is computationally more expensive and may have a limitation in detecting blurred faces. In addition, the proposed NPD method is not as good as the Soft cascade, the state-of-the-art method on the CMU-MIT dataset. Still, the proposed NPD method can detect about 80% of the frontal faces without any false positives, which is promising. Some of the detected faces in the CMU-MIT dataset by the proposed NPD method are shown in Fig. 11.

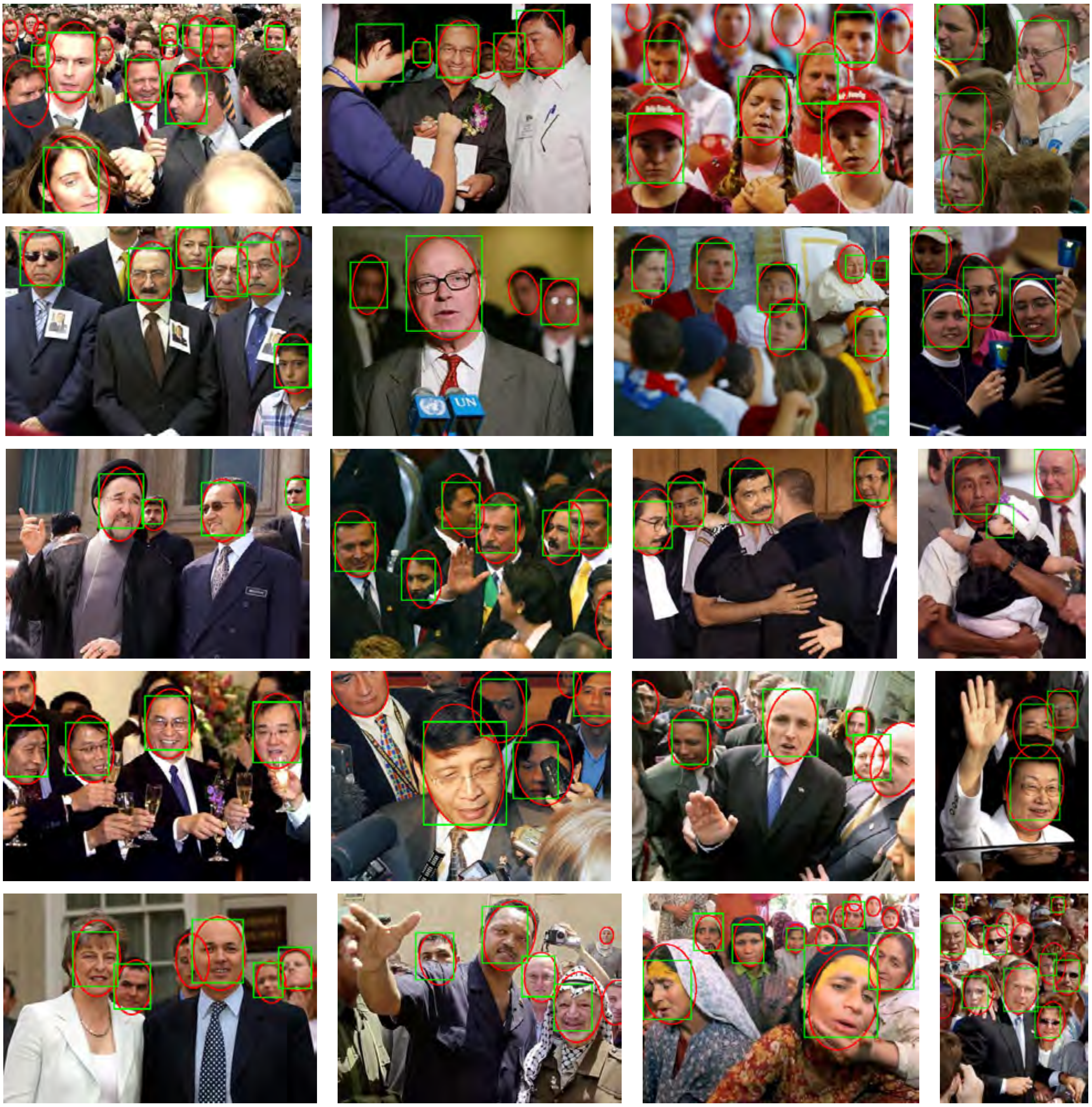

Fig. 7. Detected faces in the FDDB database [3] by the proposed NPD method. Green boxes are detections by the NPD detector, while red ellipses are ground truth annotations.

### 5.4 Analysis of the Proposed Face Detector

Since the proposed face detector is a combination of the NPD features and tree classifiers, it is instructive to determine the contribution of each of these two components. In the following, we trained all compared face detectors on the same training set [13] and cascade training settings described in Section 4.2.1.

First, we fixed the classic regression tree (CART [35]) based weak learner with depth of four, and compared the proposed NPD feature to three other local features, namely Haar-like features [1], LBP [61], and POF [26], [27], [29], [31]. Since LBP is a discrete label, we treated it as a categorical variable in the regression tree learning, that is, for branching at each tree node, the algorithm finds the optimal criterion that splits the discrete LBP codes into two groups. Using the same training set [13], we trained the three detectors using Haar, LBP, and POF, respectively. The model complexity of these detectors is summarized in Table 2. It can be observed that, the CART based NPD model is more efficient than the POF model, though it requires slightly more feature evaluations than the Haar and LBP models. However, it should be noted that the computation of Haar-like features requires computing integral images, while for LBP, each feature needs to compare 8 pairs of pixels and convert the resulting binary string to the corresponding decimal number. In contrast, using look up tables as aforementioned, computing the NPD feature requires only one memory access.

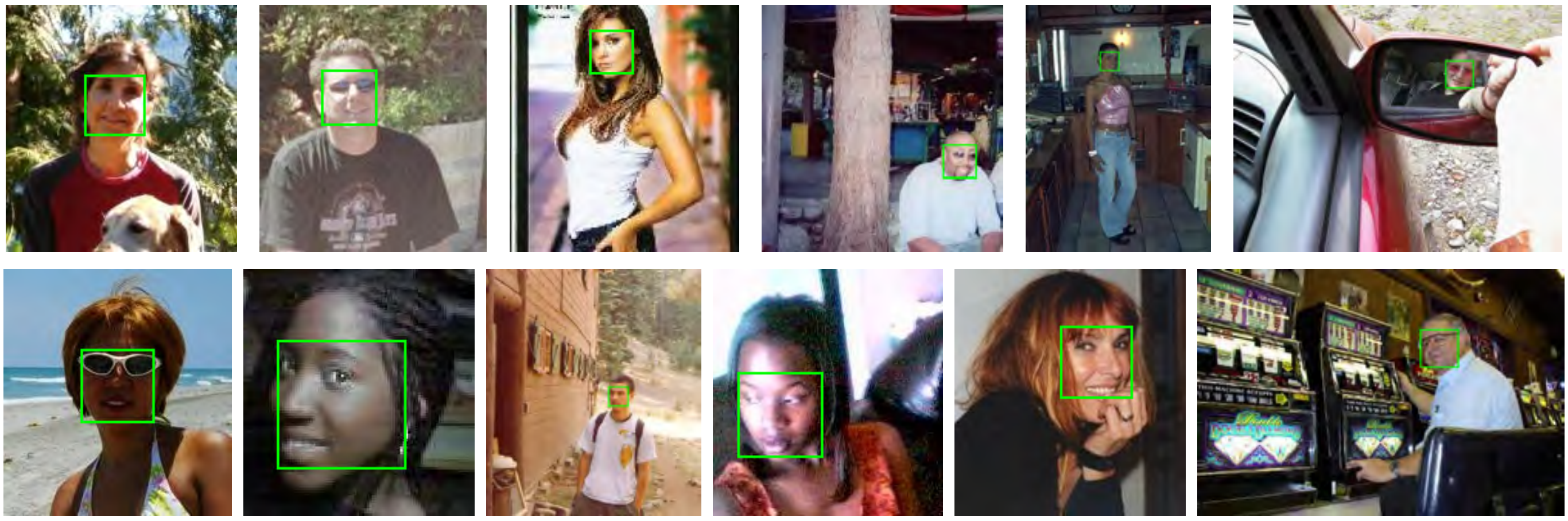

Fig. 9. Detected faces in the GENKI-SZSL dataset [54] by the proposed NPD method.

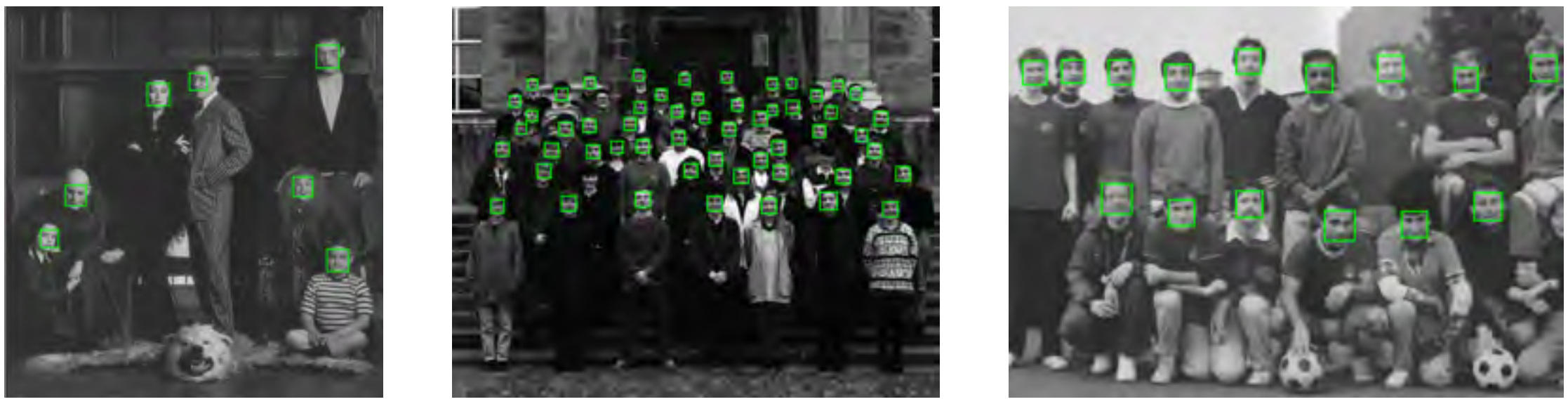

Fig. 11. Detected faces in the CMU-MIT dataset [36] by the proposed NPD method.

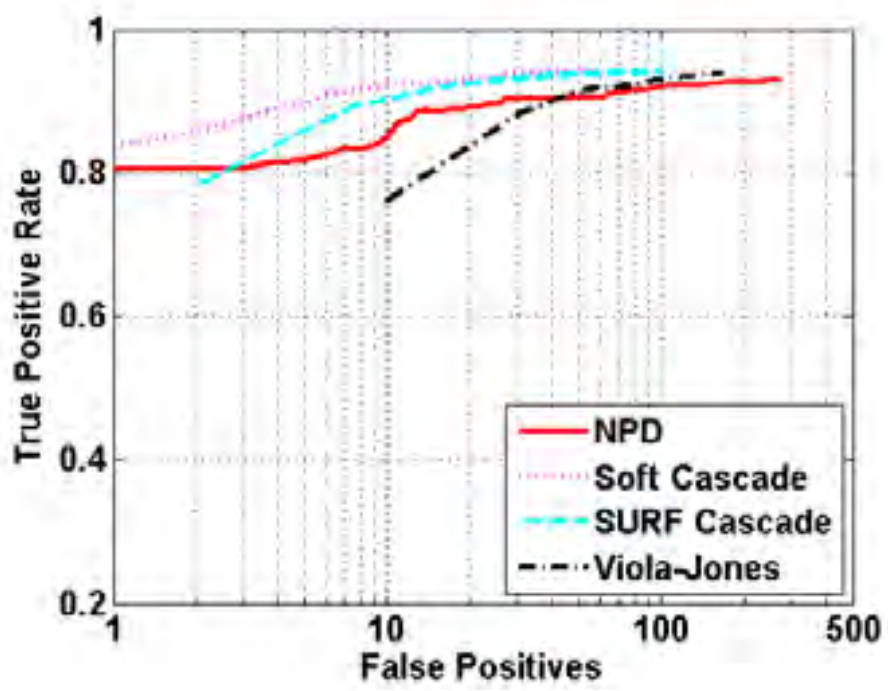

Fig. 10. ROC curves for face detection on the CMU-MIT dataset [36].

TABLE 2
Comparison of detector complexity.

| | Haar | LBP | POF | NPD | | | |
|---|---|---|---|---|---|---|---|
| | | | | Stump | CART | DQT | DQT-Soft |
| **#trees** | 150 | 108 | 276 | 1,597 | 176 | 140 | 72 |
| **#features** | 1,763 | 1,269 | 3,082 | 1,597 | 2,035 | 1,929 | 1,018 |
| **#fea. evals.** | 33.9 | 30.4 | 44.3 | 36.5 | 34.4 | 23.5 | 18.6 |

The four detectors with different local features were tested on the FDDB database, resulting in ROC curves shown in Fig. 12 for both the discrete and continuous score metrics. The NPD detector performs better than the Haar, LBP, and POF detectors with the same CART based weak learners. The performance improvements due to NPD features over Haar, LBP, and POF features are about 6%, 19%, and 15%, respectively, for discrete metric, and about 4%, 13%, and 10%, respectively, for continuous metric, at FP=1. NPD is better than POF, because with NPD features the regression tree learns optimal thresholds to form more robust ordinal rules. NPD performs better than Haar and LBP, especially at low false positives, indicating that combining optimal pixel-level features in regression trees provides better discrimination between faces and nonfaces.

We also tried a variation of NPD, defined as $f(x, y) = \frac{x-y}{\sqrt{x^2+y^2}}$ and denoted as NPD2. The comparison on FDDB are illustrated in Fig. 12, showing that the performance of NPD is slightly better than that of NPD2. Therefore, given that NPD is simpler than NPD2, we prefer the formulation of Eq. (1).

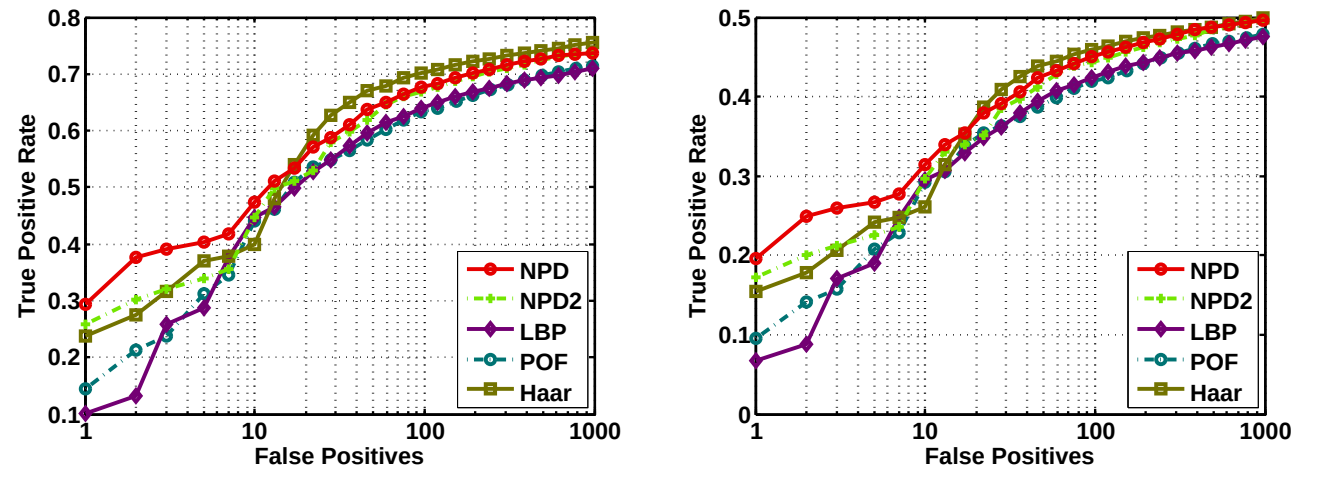

(a) discrete (b) continuous

Fig. 12. Comparison of different features in CART based face detector on the FDDB database [3] with (a) discrete and (b) continuous score metrics.

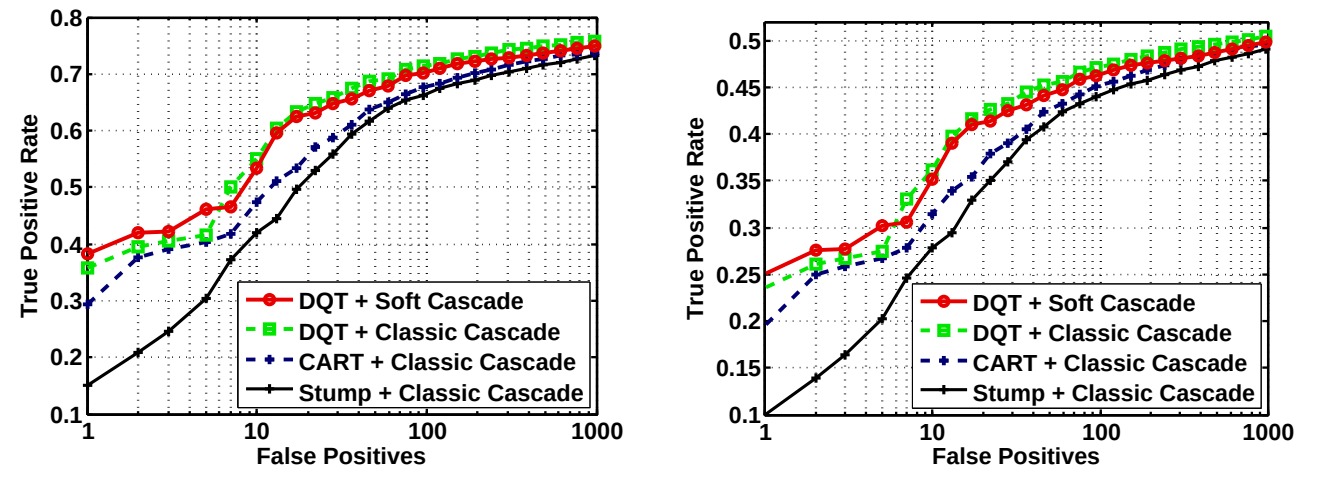

(a) discrete (b) continuous

Fig. 13. Comparison of NPD face detectors based on different weak learners on the FDDB database [3] with (a) discrete and (b) continuous score metrics.

Next, we fixed the NPD feature representation and the classic cascade architecture [1], and compared three different weak learners, namely the stump classifier [1], the classic regression tree CART, and the proposed deep quadratic tree (DQT). Both CART and DQT were with depth of four. As shown in Table 2, the stump based detector requires much more weak classifiers than CART, indicating that combining NPD features in a deeper regression tree is much more effective. Furthermore, Table 2 shows that using CART does not increase the average computation cost compared to stump w.r.t. average feature evaluations. In addition, the proposed DQT based learner further reduces the number of weak classifiers and average feature evaluations required. The three face detectors were tested on the FDDB database, resulting in ROC curves shown in Fig. 13 for both the discrete score metric and continuous score metric. As illustrated, using CART instead of stump classifier improves the face detection performance by about 0% – 17% for discrete metric and 0% – 11% for continuous metric. The improvement is larger at smaller false positives. This verifies that tree classifiers help to optimally combine NPD features for the complex unconstrained face detection task. Besides, the DQT based detector further improves the performance, due to its quadratic splitting capability compared to linear splitting.

Finally, with NPD+DQT, we compared the soft cascade detector [52] and the classic cascade detector [1], as shown in Table 2 and Fig. 13. Clearly, with comparable performance, soft cascade further reduces the model complexity.

### 5.5 Evaluation Under Specific Challenges

In the following, we evaluate how the proposed NPD face detector performs under illumination variation, pose variation, occlusion, and blur (or low resolution). Note that these four challenges are often encountered simultaneously in an image. In our selection of the four subsets, one per specific challenge, we focused on the main source of variation in each image. For each challenge, we selected 100 images from the FDDB database [3] (examples are shown in Fig. 14), and ran our unconstrained NPD face detector (trained with AFLW) on each subset separately. Fig. 15 shows that the NPD face detector performs the best on the pose and illumination subsets, thanks to the scale-invariant NPD features and the deep quadratic trees. For the occlusion and blur subsets, the performance largely drops. These results indicate that occlusion and blur are the two major challenges for unconstrained face detection, which have not been well addressed in the literature.

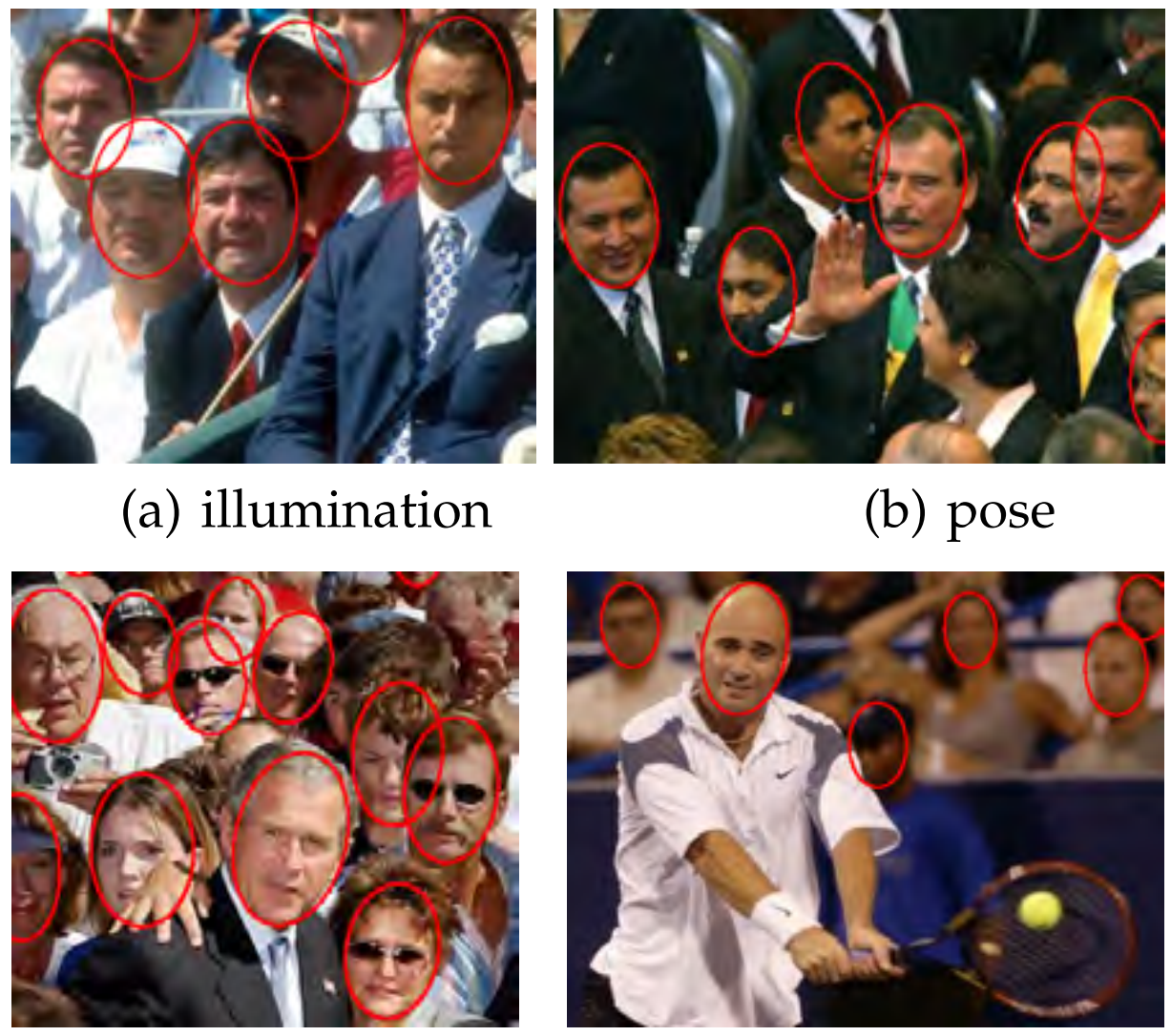

(a) illumination (b) pose (c) occlusion (d) blur

Fig. 14. Example images and annotated faces for four subsets extracted from the FDDB database [3].

The NPD face detector is also compared with the Viola-Jones face detector implemented in OpenCV 2.4, and the commercial face detector PittPatt on the four subsets of FDDB discussed above. The resulting ROC curves with the discrete score metric are shown in Fig. 16. These plots show that the proposed NPD face detector outperforms both the Viola-Jones and the PittPatt face detectors on all the four subset-

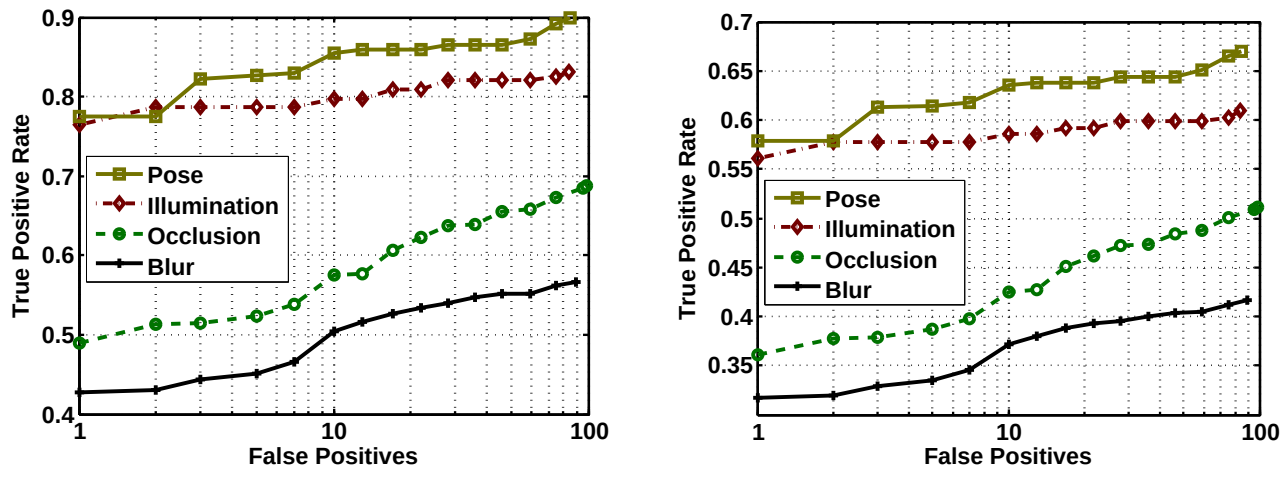

(a) discrete (b) continuous

Fig. 15. ROC curves of the proposed NPD face detector on the four subsets extracted from the FDDB database [3] with (a) discrete and (b) continuous score metrics.

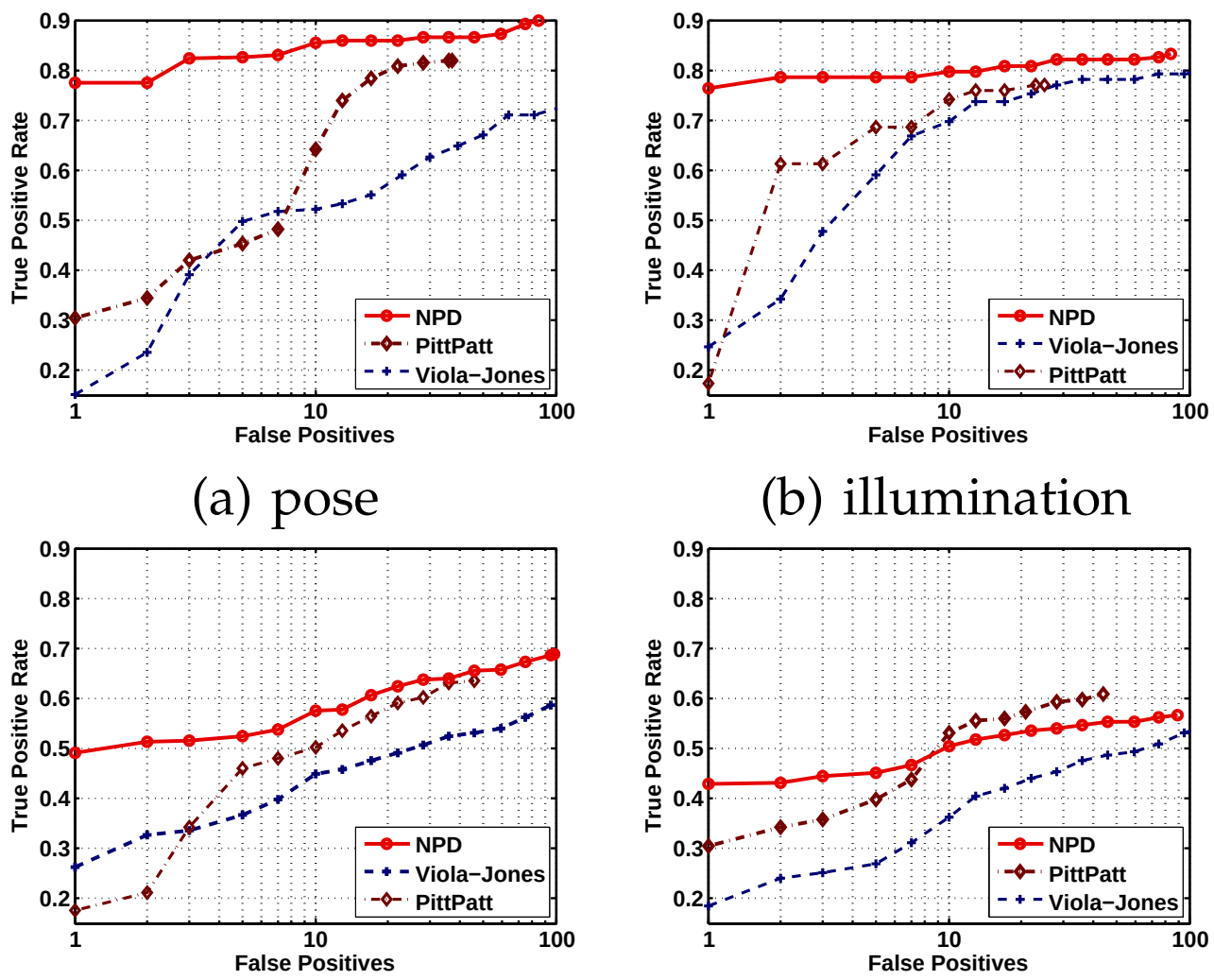

(a) pose (b) illumination

(c) occlusion (d) blur

Fig. 16. ROC curves for face detection on four subsets from the FDDB database [3] with the discrete score metric.

s. The reasons for the superior performance of the proposed method under illumination variations, pose variations, occlusions, and blur, were discussed in Subsection 4.2.

### 5.6 Detection Speed

For handheld devices like mobile phones, the available resources for computation and memory are rather limited. Therefore, face detector's complexity and detection speed are very important for embedded systems. In this subsection, we report the detection speed of the proposed NPD face detector, compared with the Viola-Jones[5] face detector in OpenCV 2.4, which is known to be optimized for speed. The proposed NPD face detector is implemented in C++, which requires about 50MB of memory in runtime. The model size of our frontal detector is 41KB, while that of the unconstrained detector is 831KB. Two platforms were selected for this evaluation: (i) a normal desktop PC with the Intel Core i5-2400 @3.1GHz CPU (4 cores, 4 threads), and (ii) a netbook with Intel Atom N450 @1.6GHz processor (1 core, 2 threads), to simulate low-end devices. For face detection evaluation, a video clip of the movie "*Jobs*" was used. This video clip shows a busy campus, with each frame containing from one to tens of faces. The length of the video clip is about 2 minutes, containing 3,950 frames in total. The original resolution is $1280 \times 720$. To test the detection speed at various resolutions, the original video clip was cropped and resized to $1920 \times 1080$, $800\times600$, and $640\times480$. In this evaluation, the minimal face size to detect was set to $40 \times 40$ pixels for frontal detector as in [13] and $80\times80$ for unconstrained detector as in [46], [48], [49], and the scaling factor was 1.2. The multi threading technique was enabled in both NPD and OpenCV detectors for parallel computation. Note that we only calculated the face detection time, regardless of the video decoding time.

The test results (in terms of Frame Per Second, FPS) of frontal detectors are shown in Table 3. The detection parameters of the SURF cascade method [13] are the same as our algorithm, except that authors in [13] used an i7@3.2GHz CPU (4 cores, 8 threads) for the desktop computer. It can be observed that the NPD detector is much faster than both the OpenCV and SURF cascade detectors. On Atom N450 processor, the detection speed of the NPD detector is about 9 times faster than the detection speed of the OpenCV detector; on i5 processor the speed of the NPD detector is about 7 times the speed of the OpenCV detector.

Table 3 shows that our frontal face detector can run in real-time (29.6 FPS) on i5 desktop PC for processing $1920 \times 1080$ high definition videos, and 177.6FPS for VGA videos. On the low-end Atom platform, the NPD detector can run in near real-time (19.4 FPS) for VGA videos. The reasons for the high processing speed of NPD are two folds. First, the NPD feature is simple, involving only two pixels. Further with the look up table technique, the evaluation of each NPD feature requires only one memory access. Second, the NPD feature can be easily scaled to various sizes of detection templates. Therefore, pre-calculating and storing multiscale templates can speed up detection because rescaling the input image is avoided.

Next, we also evaluate our unconstrained detector and compare it to recent methods, as shown in Table 4. The proposed method is much faster than Yan-DPM [46] and ACF [49] with either a single thread or multi-threads. The NPD detector achieves similar speed as that of Joint Cascade method [48]. Using multi-thread i5 CPU, we are able to achieve 70FPS for unconstrained face detection on VGA frames, which is slower than the frontal detector, but still quite efficient.

5. We have tested four models of the Viola-Jones face detector provided in OpenCV 2.4, and found that the "haarcascade_frontalface_alt" model is the fastest, which was selected here for comparison.

TABLE 3
Speed (FPS) of frontal face detectors.

| CPU | Resolution | NPD | OpenCV | SURF [13]* |
|---|---|---|---|---|
| Atom N450 | 640 × 480 | **19.4** | 2.1 | 5.8 |
| | 800 × 600 | **12.1** | 1.3 | n/a |
| @1.6GHz | 1280 × 720 | **6.8** | 0.7 | n/a |
| (1 core, 2 threads) | 1920 × 1080 | **3.0** | 0.3 | n/a |
| i5-2400 | 640 × 480 | **177.6** | 24.4 | 71.3 |
| | 800 × 600 | **112.6** | 16.2 | n/a |
| @3.1GHz | 1280 × 720 | **63.3** | 8.9 | n/a |
| (4 cores, 4 threads) | 1920 × 1080 | **29.6** | 3.6 | n/a |

* "n/a" means results are not reported in [13] for the SURF detector.

TABLE 4
Speed (FPS) of unconstrained face detectors.

| | **NPD** | | **Yan-DPM [46]** | | **JCascade [48]** | **ACF [49]** | |
|---|---|---|---|---|---|---|---|
| **CPU** | i5@3.1GHz | | X5650@2.66GHz | | @2.93GHz | i7@3.9GHz | |
| **Cores** | 4 | | 6 | | n/a | 4 | |
| **Threads** | 1 | 4 | 1 | 12 | 1 | 1 | 8 |
| **Speed** | 29.28 | 70.06 | 5 | 25 | 34.97 | 15 | 42 |

## 6 Summary and Future Work

We have proposed a fast and accurate method for face detection in cluttered scenes. First, a simple feature called NPD is proposed, which has properties of scale invariance, boundedness, and reconstruction ability. Second, we propose a deep quadratic tree to learn the optimal subset of NPD features and their combinations. As a result, a single soft-cascade AdaBoost classifier is able to achieve promising results for face detection with large pose variations and occlusions. Evaluations on three public face databases show that the proposed method achieves state-of-the-art performance for unconstrained face detection, and an analysis show that occlusions and blur are two big challenges for face detection. The proposed detector is also efficient, about 6 times faster than the Viola-Jones face detector implemented in OpenCV 2.4. It is interesting to apply the proposed NPD feature and the classifier learning method for other tasks such as face attribute classification and pedestrian detection.

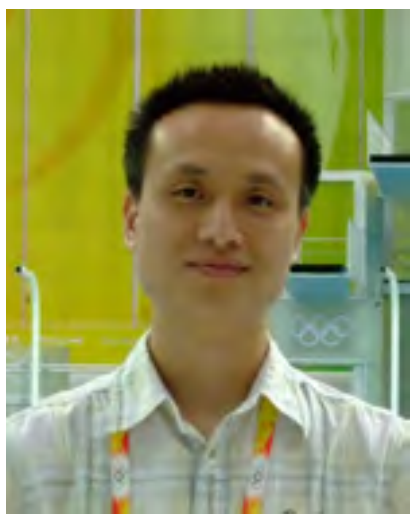

**Shengcai Liao** received the B.S. degree in mathematics and applied mathematics from the Sun Yat-sen University in 2005, and the Ph.D. degree from the Institute of Automation, Chinese Academy of Sciences (CASIA) in 2010. He was a Postdoc in the Dept. of Computer Science and Engineering, Michigan State University during 2010-2012, and he is now an Associate Professor in CASIA. His research interests include face recognition and video surveillance. He was awarded the Excellence Paper of Motorola Best Student Paper and the 1st Place Best Biometrics Paper award at the International Conference on Biometrics in 2006 and 2007, respectively. He was also awarded the best reviewer award in IJCB 2014.

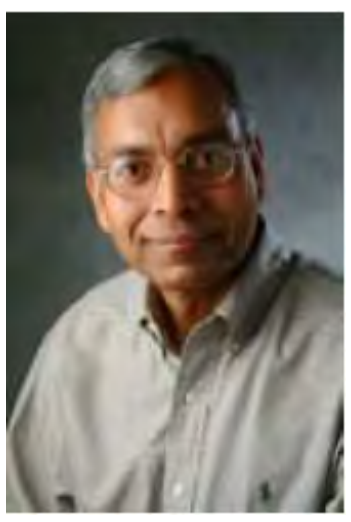

**Anil K. Jain** is a university distinguished professor in the Department of Computer Science and Engineering at Michigan State University. His research interests include pattern recognition and biometric authentication. He served as the editor-in-chief of the IEEE TPAMI (1991-1994). The holder of six patents in the area of fingerprints, he is the author of a number of books, including Handbook of Fingerprint Recognition (2009), Handbook of Biometrics (2011), Handbook of Multibiometrics (2006), Handbook of Face Recognition (2005), BIOMETRICS: Personal Identification in Networked Society (1999), and Algorithms for Clustering Data (1988). He served as a member of the Defense Science Board and The National Academies committees on Whither Biometrics and Improvised Explosive Devices. Dr. Jain received the 1996 IEEE TNN Outstanding Paper Award and the Pattern Recognition Society best paper awards in 1987, 1991, and 2005. He is a fellow of the AAAS, ACM, IAPR, and SPIE. He has received Fulbright, Guggenheim, Alexander von Humboldt, IEEE Computer Society Technical Achievement, IEEE Wallace McDowell, ICDM Research Contributions, and IAPR King-Sun Fu awards.

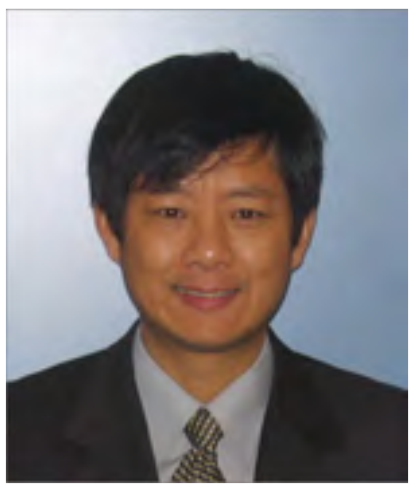

**Stan Z. Li** received the B.Eng. degree from Hunan University, Changsha, China, the M.Eng. degree from the National University of Defense Technology, China, and the Ph.D. degree from Surrey University, Surrey, U.K. He is currently a Professor and the Director of Center for Biometrics and Security Research (CBSR), CASIA. He worked at Microsoft Research Asia as a researcher from 2000 to 2004. Prior to that, he was an Associate Professor at Nanyang Technological University, Singapore. His research interest includes pattern recognition and machine learning, image and vision processing, face recognition, biometrics, and intelligent video surveillance. He has published over 200 papers in international journals and conferences, and authored and edited eight books. Dr. Li was an Associate Editor of the IEEE TPAMI and is acting as the Editor-in-Chief for the Encyclopedia of Biometrics. He served as a co-chair for the International Conference on Biometrics 2007 and 2009, and has been involved in organizing other international conferences and workshops in the fields of his research interest.

# Appendix A
# Boundedness of NPD

***Lemma 1 (Boundedness):*** $\forall x, y \geq 0$, the NPD feature f(x,y) is well bounded in [-1,1]. In addition, $f(x, y) = 1$ if and only if $x > 0$ and $y = 0$; and $f(x, y) = -1$ if and only if $x = 0$ and $y > 0$.

**Proof:** From the definition of NPD we know that $x \geq 0$, $y \geq 0$, and $f(0, 0) = 0 \in [-1, 1]$. When either $x$ or $y$ is nonzero, for example, $y \geq 0$ but $x > 0$, Eq. (1) can be reformulated as

$$f(x, y) = \frac{x - y}{x + y} = \frac{2x}{x + y} - 1 = \frac{2}{1 + \frac{y}{x}} - 1 \leq 1. \quad \text{(a)}$$

The inequality in Eq. (a) holds because $y \geq 0$, and the last equality holds if and only if $x > 0$ and $y = 0$. Similarly, when $x \geq 0$ but $y > 0$, Eq. (1) can be reformulated as

$$f(x, y) = \frac{x - y}{x + y} = 1 - \frac{2y}{x + y} = 1 - \frac{2}{\frac{x}{y} + 1} \geq -1. \quad \text{(b)}$$

The inequality in Eq. (b) holds because $x \geq 0$, and the last equality holds if and only if $x = 0$ and $y > 0$. □

# Appendix B
# Proof of Theorem 1

Denote $f_{ij} = f(x_i, x_j)$. From Eq. (1) we have

$$f_{ij}(x_i + x_j) = x_i - x_j. \quad \text{(c)}$$

Equivalently,

$$(f_{ij} - 1)x_i + (f_{ij} + 1)x_j = 0. \quad \text{(d)}$$

Therefore, we have the following set of linear equations

$$\mathbf{F}\mathbf{x} = \mathbf{0}, \quad \text{(e)}$$

where

$$\mathbf{F} = \begin{pmatrix} f_{12} - 1 & f_{12} + 1 & 0 & \cdots & 0 \\ f_{13} - 1 & 0 & f_{13} + 1 & \cdots & 0 \\ \cdots & \cdots & \cdots & \cdots & \cdots \\ f_{1p} - 1 & 0 & 0 & \cdots & f_{1p} + 1 \\ 0 & f_{23} - 1 & f_{23} + 1 & \cdots & 0 \\ \cdots & \cdots & \cdots & \cdots & \cdots \\ 0 & 0 & 0 & \cdots & f_{p-1,p} + 1 \end{pmatrix} \quad \text{(f)}$$

is a sparse $d \times p$ matrix with each row containing at most two nonzero entries. Furthermore, from the formulation of $\mathbf{F}$ we know that each row of $\mathbf{F}$ contains at least one nonzero entry, because $(f_{ij} - 1) \neq (f_{ij} + 1)$ always holds for all $i$ and $j$. Without loss of generality, let's assume $f_{12} + 1 \neq 0$. Then it follows that $f_{1j} + 1 \neq 0, \forall j$. Because if $\exists j$ such that $f_{1j} + 1 = 0$, then from Lemma 1 we know that $x_1 = 0$. This will further lead to $f_{12} + 1 = 0$, which violates the assumption that $f_{12} + 1 \neq 0$. Therefore, the first $p - 1$ rows in the matrix $\mathbf{F}$ are linearly independent of each other.

We will further prove that $rank(\mathbf{F}) = p - 1$. In fact, any row of the matrix $\mathbf{F}$ can be linearly expressed by the first $p - 1$ rows. To show this, let's denote the row containing $f_{ij} - 1$ and $f_{ij} + 1$ by $\mathbf{r}_{ij}$. We will show that

$$\mathbf{r}_{ij} = \frac{f_{ij} - 1}{f_{1i} + 1}\mathbf{r}_{1i} + \frac{f_{ij} + 1}{f_{1j} + 1}\mathbf{r}_{1j}, \quad \text{(g)}$$

holds for all $i > 1$ and $j > i$. In fact, it is easy to verify that the above equation holds for all columns of $\mathbf{r}_{ij}$, $\mathbf{r}_{1i}$, and $\mathbf{r}_{1j}$ after the first column. So, we only need to show that, for the first column, we have

$$\frac{(f_{1i} - 1)(f_{ij} - 1)}{f_{1i} + 1} + \frac{(f_{1j} - 1)(f_{ij} + 1)}{f_{1j} + 1} = 0, \quad \text{(h)}$$

which is equivalent to

$$f_{1i}f_{1j}f_{ij} - f_{1i} + f_{1j} - f_{ij} = 0. \quad \text{(i)}$$

This can be verified by substituting each feature with its definition in Eq. (1).

Given that $rank(\mathbf{F}) = p - 1$, we know that the nullspace of $\mathbf{F}$ contains only one nonzero vector, which is a solution to Eq. (e). Furthermore, from Lemma 1 we can infer that $(f_{ij} - 1)(f_{ij} + 1) \leq 0$, hence Eq. (d) tells that $x_i x_j \geq 0, \forall i, j$. Consequently, Eq. (e) always has a nonnegative solution $\hat{\mathbf{x}}$, and all solutions to Eq. (e) must be $c\hat{\mathbf{x}}$, where $c$ is a scale factor. □

Given this proof, we make four observations below:

- For a solution, $c$ can be any real value, but to satisfy the constraint that all pixel intensity values are nonnegative, $c$ should be positive.
- The solution to Eq. (e) spans a one-dimensional subspace (the nullspace).
- A specific solution can be obtained by assigning $x_1 = 1$ and solving for the other variables from the first $p - 1$ rows of Eq. (e) in linear time.
- When the original image is $\mathbf{x} = \mathbf{0}$, it can also be reconstructed by $c\hat{\mathbf{x}}$ where $\hat{\mathbf{x}}_i = 1, \forall i$, and $c = 0$. However, in this case a solution with $c > 0$ is not generally regarded as a scaled version of the original image $\mathbf{x} = \mathbf{0}$.